\documentclass[letterpaper, 10 pt, journal, twoside]{IEEEtran}

\usepackage{graphicx}
\IEEEoverridecommandlockouts

\usepackage{amsmath}
\usepackage{amssymb}
\usepackage{mathtools}
\usepackage{bm}
\usepackage{lipsum}
\usepackage{graphicx}
\usepackage{algorithm}
\usepackage{algpseudocode}
\usepackage{tabularx}
\usepackage{array}
\usepackage{gensymb}
\usepackage{booktabs}
\usepackage{stfloats}

\makeatletter
\let\NAT@parse\undefined
\makeatother
\usepackage[hidelinks]{hyperref}
\usepackage{tikz}
\usetikzlibrary{shapes,arrows.meta,positioning}

\newcommand{\Rc}{\mathcal R}
\title{Fault-Tolerant, Rigidity-Preserving Control of Inflatable Truss Robots}

\author{
    James Wade, Isaac Weaver, Mihai Stanciu, Nathan Usevitch
    \thanks{Manuscript received: May 15, 2026; Revised August 10, 2026; Accepted September 1, 2026. This paper was recommended for publication by Editor B. Mazzolai upon evaluation of the Associate Editor and Reviewers’ comments. This work was supported in part by the National Science Foundation under Grant 2501928 and in part through the Utah NASA Space Grant Consortium, NASA Grant \#80NSSC25M7088.}
    \thanks{The authors are with the Ira A. Fulton College of Engineering, Mechanical Engineering Department, Brigham Young University, Provo, Utah, USA (e-mail: {\tt\footnotesize note2jwade@gmail.com}, {\tt\footnotesize nathan\_usevitch@byu.edu)}.}
    \thanks{A supplementary video is available at \url{https://youtu.be/T30hJi9g9XE}.}
    \thanks{Digital Object Identifier (DOI): 10.1109/LRA.2026.3733614}
}

\usepackage{fancyhdr}

\fancypagestyle{copyrightpage}{
    \fancyhf{}
    \fancyfoot[C]{%
        \parbox{0.95\textwidth}{%
            \centering\scriptsize
            \textcopyright~2026 IEEE. All rights reserved, including rights for
            text and data mining and training of artificial intelligence and
            similar technologies. Personal use is permitted, but
            republication/redistribution requires IEEE permission.
            See https://www.ieee.org/publications/rights/index.html for more information.
        }%
    }

}

\begin{document}
\markboth{IEEE Robotics and Automation Letters. Preprint Version. Accepted September, 2026}
{Wade \MakeLowercase{\textit{et al.}}: Fault-Tolerant, Rigidity-Preserving Control of Inflatable Truss Robots} 
\maketitle
\thispagestyle{copyrightpage}

\begin{abstract}
Isoperimetric robotic trusses can adapt to different tasks and environments due to their high strength-to-weight ratio and the ability to reconfigure dramatically into new shapes. However, motor failures in operational environments can severely limit capability if not properly addressed. This paper presents a fault-tolerant control framework for an inflatable robotic truss that maintains functionality despite motor failures, demonstrated through three key contributions. First, we extend the kinematic optimization to handle arbitrary combinations of motor failures by imposing equality constraints to ensure failed actuators are not used. Second, we introduce discrete-time control barrier function (DTCBF) constraints that, under the assumed kinematic model, mathematically guarantee structural rigidity while maximizing workspace utilization. Third, we implement closed-loop position control using onboard encoder feedback and a forward-kinematics-based state estimator, improving positional accuracy in the presence of disturbances. We validate our approach through simulation and hardware experiments on a 2D isoperimetric truss testbed. For a 2D configuration with 6 actuators, we demonstrate $\mathbf{>62\%}$ workspace preservation under single-motor failures, a $\mathbf{>22\%}$ reduction in tracking RMSE using failure-aware closed-loop control relative to the fully functional open-loop baseline, and a $\mathbf{39\%}$ closed-loop RMSE reduction over failure-unaware open-loop control under unmodeled partial actuator degradation. These results establish a foundation for more robust and resilient isoperimetric truss robots operating under degraded actuation.
\end{abstract}


\section{Introduction}
\begin{figure}[tbp]
  \centering
  \includegraphics[width=\linewidth]{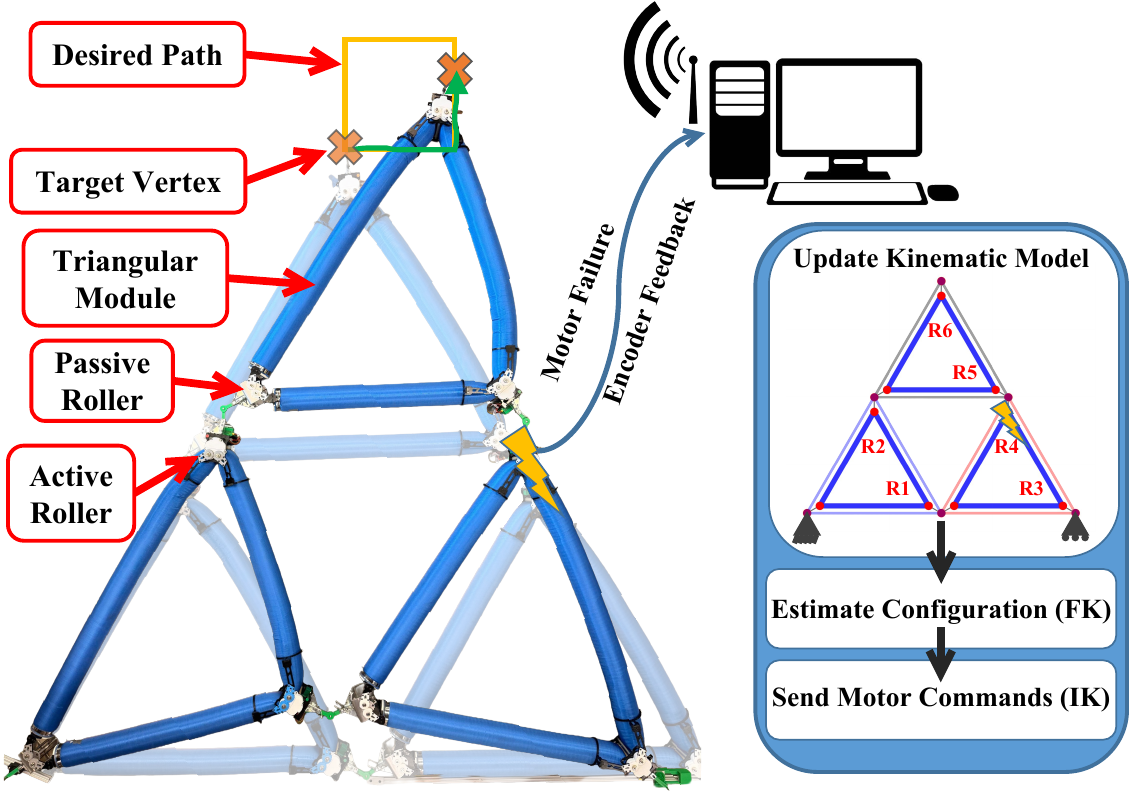}
  \caption{
  The 2D isoperimetric truss robot testbed illustrating the closed-loop, failure-aware control architecture. The target vertex traces a square path (yellow); the translucent overlay shows a prior configuration. A lightning bolt marks a motor failure, communicated with encoder feedback to the central computer, which updates the kinematic model, estimates the current configuration via forward kinematics (FK), and issues compensating motor commands via inverse kinematics (IK).\vspace{-2ex}}
  \label{fig:system_diagram}
\end{figure}

\IEEEPARstart{R}{econfigurable} truss robots offer a lightweight, deployable, and adaptable framework for robotic structures. Their high deployed-to-stowed volume ratio and ability to morph between configurations suit applications such as remote infrastructure deployment, field robotics, and space systems \cite{stanciu2026modular}. This work focuses on isoperimetric truss robots, which consist of compliant beams routed through motorized rollers that reconfigure the structure by redistributing tube length while maintaining a constant overall perimeter. However, existing control architectures offer limited robustness to actuator failures despite the inherent redundancy of these systems. In tightly coupled truss robots, a single actuator failure can substantially degrade workspace, manipulability, and tracking performance in ways that remain uncharacterized. This work develops a fault-tolerant control framework for inflatable isoperimetric truss robots that emphasizes maintaining rigidity and task-space performance under degraded actuation. Fig.~\ref{fig:system_diagram} illustrates the hardware platform and control architecture.

\section{System Overview}
\subsection{Isoperimetric Robots}

\begin{figure}[tbp]
  \centering
  \includegraphics[width=.98\linewidth]{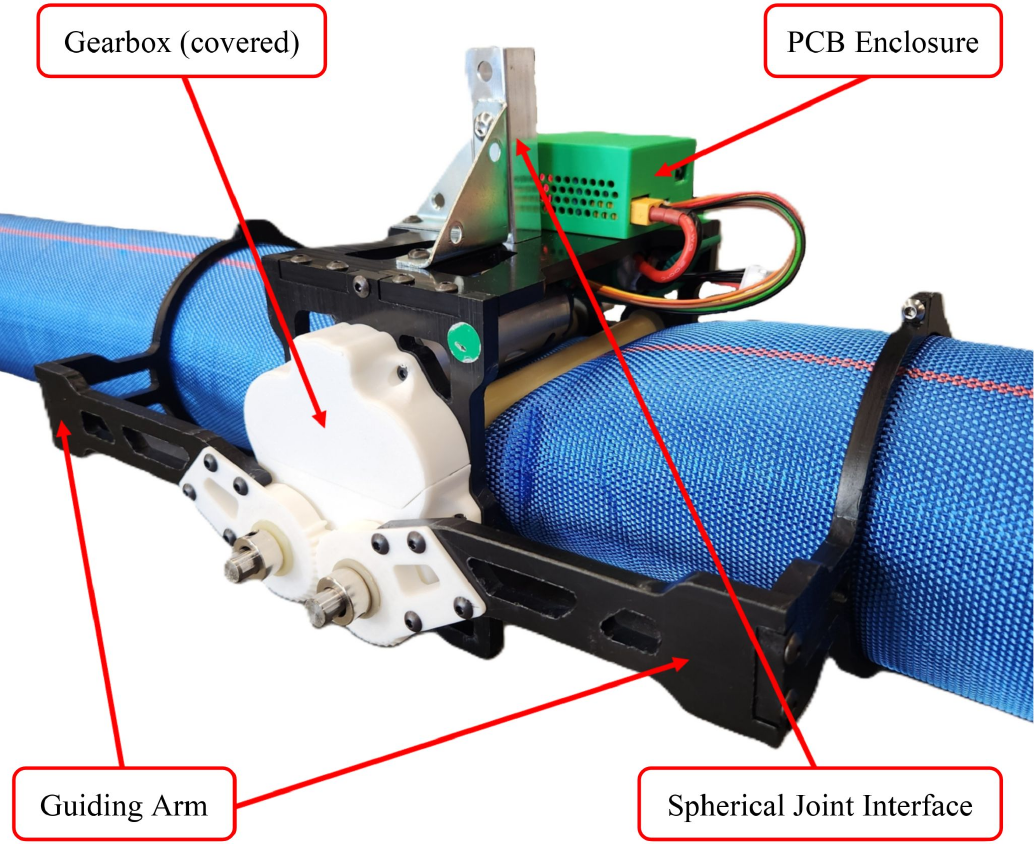}
  \caption{Diagram labeling each part of the truss robot's active roller unit. Three roller nodes and an inflated tube form a triangular module of the truss robot.}
  \label{fig:active_roller}
  \vspace{-4.5mm}
\end{figure}

Isoperimetric robots were first introduced in \cite{14_usevitch_untethered} and later extended as modular truss-like structures for lunar environments \cite{stanciu2026modular}. The robot is constructed from triangular modules, each consisting of a single inflated tube routed through three roller nodes: two active roller units (Fig.~\ref{fig:active_roller}) and one passive roller. Each active roller contains a DC motor that redistributes tubing between triangle edges, changing the triangle's shape while its perimeter remains constant by construction. The passive roller shares the same mechanical structure but omits the motor and PCB, serving to anchor the bent tube ends. This inflatable truss architecture combines lightweight transportability, adaptability, and a high strength-to-weight ratio \cite{stanciu2026modular}, offering a promising alternative to rigid robots. However, the tightly coupled actuation means actuator failures can significantly degrade mobility, motivating fault-sensitivity analysis.

\subsection{Limitations of Current Isoperimetric Robot Controller}
A key capability of the robot is moving a user-specified target vertex (Fig.~\ref{fig:system_diagram}) along a desired trajectory. The previous kinematics-based controller solved a quadratic optimization problem for each active roller motor's motion, assuming all actuators are fully functional \cite{usevitch2025triangledecomposable}. In practice, motor failures and imperfections can significantly reduce task performance if the controller cannot adapt to the degraded system state, thus motivating an adaptive optimization-based controller.

\subsection{Proposed Solution and Contributions}
To address these limitations, we developed a robust controller for the isoperimetric robot capable of operating under arbitrary combinations of motor failures. The controller maximizes trajectory accuracy and usable workspace despite reduced actuation capability, while discrete-time control barrier functions enable safe navigation along workspace boundaries. The resulting system is more resilient to actuator failures and better suited for reliability-critical operations. This work makes the following contributions:

\begin{itemize}
    \item Robustness analysis of the motion of an isoperimetric robot under motor failures, including workspace and manipulability degradation.
    \item Discrete-time control barrier function formulation providing a model-based guarantee of structural rigidity while maximizing workspace utilization.
    \item Closed-loop position control using onboard sensing and forward-kinematics-based state estimation for an inflatable isoperimetric robot.
\end{itemize}

\section{Related Work}
\paragraph{Truss-based Robotic Systems} Truss-based robotic and structural systems have been widely explored for their high strength-to-weight ratio and efficient load-bearing capabilities. NASA has researched trusses made from lunar regolith \cite{6_nunan_lunar_truss} and trusses composed of deployable beams \cite{rhodes1985deployable} for extraterrestrial operation. Actuated truss structures in the literature include variable geometry trusses (VGTs) \cite{MIURA1985599, VTT}, tensegrity-based robots \cite{tensegrity}, and 3D-printed tetrahedral robots \cite{qin2022trussbot}. While these systems provide mechanical redundancy, they operate under fundamentally different kinematic constraints than our isoperimetric system's constant-perimeter requirement, necessitating a distinct approach to fault-tolerant control. Our truss robot is distinguished by its compliant, reconfigurable structure, enabling it to dynamically change shape and adapt to a broad range of applications. Unlike most existing truss robots, it achieves shape change by redistributing its internal compliant tubes rather than using linear actuators or tensioned strings.

\paragraph{Control of Redundant and Coupled Robotic Systems} Redundancy has long been central to robotics \cite{nenchev1989redundancy}, with much of this work focusing on resolving kinematic redundancy to satisfy secondary objectives such as singularity avoidance or fault tolerance \cite{siciliano1990kinematic, lewis1997fault}. Prior work has extensively examined the importance of maintaining graph rigidity in both theory and practice, often within the context of swarm robotics \cite{Krick01032009, autonomous_formations}. This has included motion planning for load manipulation tasks using swarms \cite{swarm_motion_planning}, similar to the load-bearing capabilities of our truss robot \cite{stanciu2026modular}. However, our approach differs fundamentally: whereas swarm agents move semi-independently, our truss robot exhibits tightly coupled motion, where displacing a single vertex induces correlated motion across the entire structure. This behavior is more analogous to hyper-redundant and continuum architectures \cite{chirikjian1994modal, jones2006kinematics}, where feasible motion emerges from coordinated whole-body deformation rather than independent agent motion. This intrinsic interdependence means that graph-based swarm methods do not directly transfer to our system, motivating extensions to the kinematic and control framework tailored to tightly coupled truss structures. In our framework, this redundancy is leveraged not only for workspace expansion, but also to satisfy rigidity-preserving safety constraints through coordinated whole-body motion.

\paragraph{Control Barrier Functions} The operational necessity of workspace utilization must be balanced with guarantees of rigidity preservation. Control barrier functions (CBFs) enable safety-critical control \cite{ames2017control} by minimally modifying the control input to maintain a nonnegative function $h(x)$ defining a safe region. For sampled-data systems such as ours, discrete-time control barrier functions (DTCBFs) provide rigorous safety guarantees at each sampling instant \cite{agrawal2017discrete, zeng2021safety}. Recent work has established sufficient conditions for forward invariance in discrete-time systems \cite{buch2023discrete}, which we leverage to ensure structural rigidity. However, DTCBFs have not previously been applied to rigidity maintenance in robotic truss systems, where redundant actuation can help maximize workspace utilization while satisfying rigidity-preserving safety constraints.

\section{Methods}
\begin{figure}
    \centering
    \includegraphics[width=\linewidth]{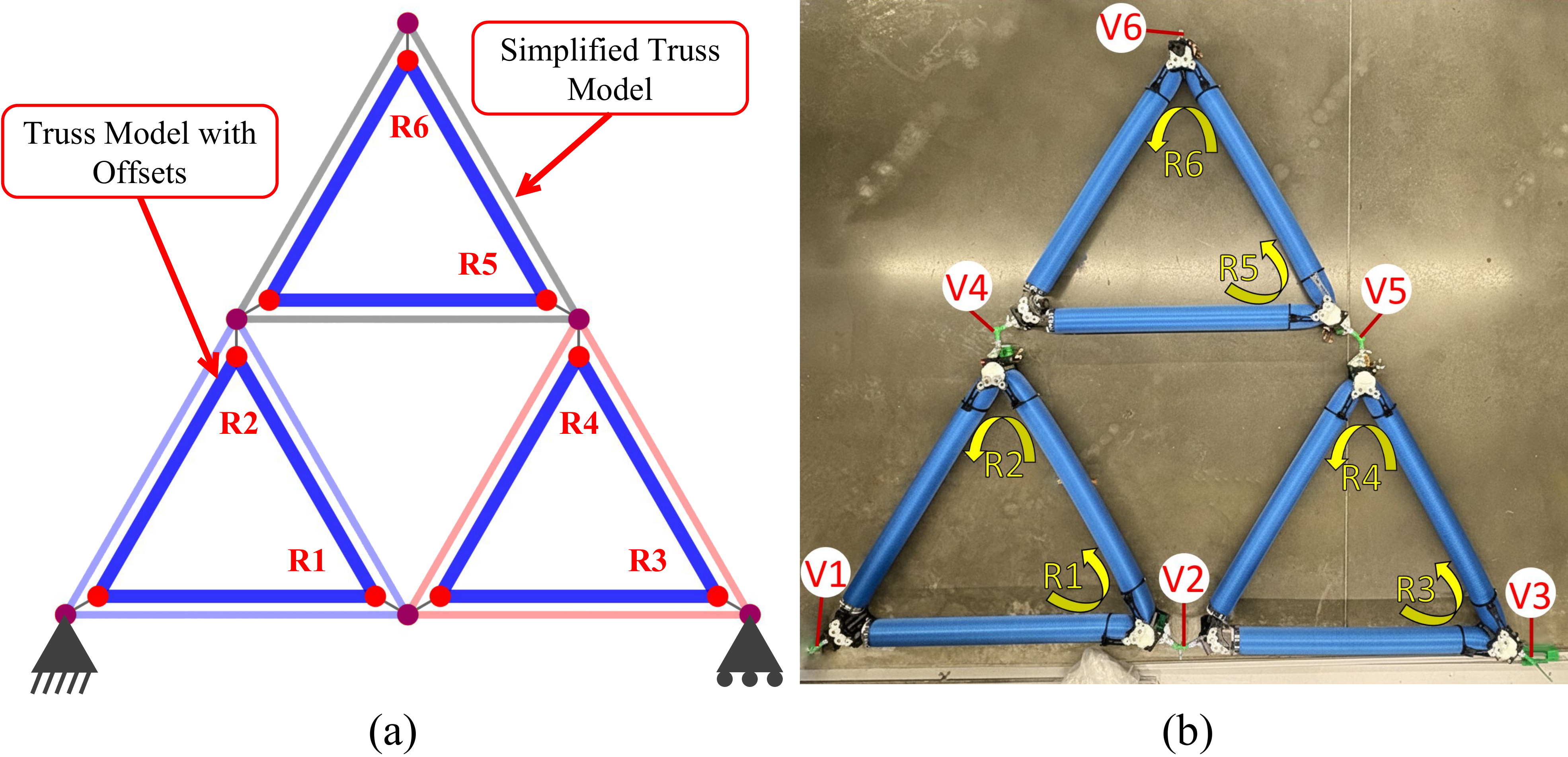}
    \caption{Two-dimensional truss robot testbed. (a) Kinematic model showing the simplified truss structure (translucent triangles) and the offset model accounting for roller geometry (blue triangles), with active rollers R1--R6 labeled at each edge. (b) Physical hardware testbed with vertices V1--V6 and rollers R1--R6 labeled; yellow arrows indicate positive roller motion.}
    \label{fig:2D_robot_diagram}
\end{figure}

This section reviews the kinematic framework of \cite{14_usevitch_untethered} and presents our novel contributions: fault-tolerant actuation, rigidity preservation, and closed-loop position control. Fig.~\ref{fig:2D_robot_diagram} shows the 2D truss robot's kinematic model and hardware setup.

\subsection{Review of Actuated System Model of the Truss Robot}
\label{sec:kinematics}

We model the truss as a graph $G = (V, E)$ in $D$-dimensional space, where $V = \{1, \ldots, N\}$ is the set of vertices with positions $p_i \in \mathbb{R}^D$, and $E$ contains $N_L$ undirected edges. The configuration and velocity vectors are $x = [p_1^\top, \ldots, p_N^\top]^\top$ and $\dot{x} = [\dot{p}_1^\top, \ldots, \dot{p}_N^\top]^\top$, respectively.

Edge lengths and their rates of change are given by
\begin{equation}\label{eq:edge_length}
    L_e = \|p_i - p_j\|, \qquad \dot{L}_e = \frac{(p_i-p_j)^\top}{\|p_i-p_j\|}\dot{p}_i + \frac{(p_j-p_i)^\top}{\|p_i-p_j\|}\dot{p}_j,
\end{equation}
yielding the well-known rigidity matrix relation \cite{14_usevitch_untethered, zelazo2012rigidity}
\begin{equation}\label{eq:rigidity}
    \dot{L} = \mathcal{R}(x)\dot{x}.
\end{equation}

The $N_{\text{roller}}$ DC motor-driven rollers actuate the structure by translating along their tubes. The rate of change of each edge in $G$ depends on the velocities of the two rollers at its endpoints,

\begin{equation}\label{eq:Ldot_rollers}
    \dot{L}_e = \pm\dot{d}_i \pm \dot{d}_j \qquad \forall e=\{i,j\} \in E,
\end{equation}
which can be written compactly as $\dot{L} = B^\top \dot{d}$, where $B \in \mathbb{R}^{N_{\text{roller}} \times N_L}$ is a sparse signed matrix whose nonzero entries are $\pm 1$, indicating each roller’s contribution to each edge-length rate and direction. Combining with Eq.~\ref{eq:rigidity} and taking the left pseudoinverse of $B^\top$ yields the inverse kinematics
\begin{equation}\label{eq:IK}
    \dot{d} = (B^\top)^\dagger \mathcal{R}(x)\dot{x}.
\end{equation}

Fig.~\ref{fig:2D_robot_diagram} shows two kinematic models: a simplified model, where each roller coincides with its vertex, and an offset model, which captures the displacement between the roller contact point and joint center. We use the offset model for state estimation and the simplified model for kinematic analysis, as it generalizes beyond the isoperimetric case. The offset model is incorporated by augmenting $G$ via the methods in \cite{14_usevitch_untethered}.

\subsection{Baseline Optimization Framework}
Summarizing the framework established in \cite{14_usevitch_untethered}, we determine optimal vertex velocities $\dot{x}^*$ by solving a baseline quadratic program (QP). This formulation handles task-space objectives through a set of linear equality constraints:
\begin{itemize}
    \item \textbf{Motion Constraints ($\mathbf{A_{\text{move}}\dot{x} = b_{\text{move}}}$):} Enforce desired task-space velocities for the user-defined target vertex.
    \item \textbf{Rigid-Body Constraints ($\mathbf{A_{\text{fixed}}\dot{x} = 0}$):} Remove global translations and rotations (6 DoF in 3D, 3 DoF in 2D) to ensure a well-posed optimization.
    \item \textbf{Loop-Closure Constraints ($\mathbf{A_{\text{loop}}\dot{x} = 0}$):} Enforce the perimeter within each triangular module by ensuring the sum of edge-length rates $\dot{L}$ for each triangle is zero. 
\end{itemize}
These constraints are included in Eq.~\ref{eq:qp_new}. The baseline objective minimizes the instantaneous rate of change of the edge lengths, augmented with a formation-preserving term:
\begin{equation}
\min_{\dot{x}} \; \underbrace{\|\Rc \dot{x}\|^2}_{\|\dot{L}\|^2} + \underbrace{k_f (\Delta L)^\top \Rc \dot{x}}_{\text{formation preservation}},
\label{eq:baseline_obj}
\end{equation}
where $\Delta L = L(x) - L_0$ denotes the deviation from the nominal edge lengths at the current timestep, and $k_f$ is a tunable scalar hyperparameter \cite{usevitch2017linear}. Eq.~\ref{eq:baseline_obj} balances efficiency of motion with proximity to the nominal configuration. Empirically, this regularization encourages the robot to return to well-conditioned configurations while the optimizer remains primarily focused on tracking the target vertex.

\subsection{Fault-Tolerant Actuation Constraints} \label{sec:broken_kinematics}
We now extend the baseline framework for fault tolerance. A failed roller module could be detected via encoder feedback when it fails to respond to commanded motions within a specified time interval. In this work, the failure set is instead directly injected with ground-truth knowledge, leaving online detection to future, hardware-specific work. If a roller motor fails, we must explicitly enforce $\dot{d}_i = 0$ for each failed roller $i$. Let $\mathcal{F} \subseteq \{1,\dots,N_{\text{roller}}\}$ denote the set of failed roller indices. Using Eq.~\ref{eq:IK}, this condition yields, for each $i \in \mathcal{F}$, the linear relation
\begin{equation}\label{eq:broken}
    \dot{d}_i = \big[(B^\top)^{\dagger}\mathcal{R}(x)\big]_{i,:}\,\dot{x} = 0,
\end{equation}
where $[\cdot]_{i,:}$ denotes the $i$th row. Stacking Eq.~\ref{eq:broken} over all $i \in \mathcal{F}$ yields the equality constraint $A_{\text{broken}}\dot{x} = 0$, where $A_{\text{broken}}$ is formed from the rows of $(B^\top)^{\dagger}\mathcal{R}(x)$ corresponding to the failed rollers, generalizing to any combination of failures.

\subsection{Rigidity Preservation via Discrete-Time Barrier Functions}

A key challenge for the truss robot is maximizing workspace while preserving structural rigidity. For infinitesimally rigid graphs, the rigidity matrix $\Rc(x)$ has exactly six zero-valued singular values in 3D (three in 2D) \cite{zelazo2012rigidity}. Sorting the singular values in ascending order, the next smallest singular value, $\sigma_7(\Rc(x))$ in 3D (or $\sigma_4$ in 2D), measures proximity to rigidity loss: near zero, the structure approaches singularity, and the risk of collapse increases \cite{zelazo2012rigidity}. We denote this value as $\sigma_{\text{crit}}$.

Rather than conservatively halting motion near singularities, we enforce rigidity via a discrete-time control barrier function (DTCBF) \cite{agrawal2017discrete,zeng2021safety}. Because the controller receives encoder feedback at discrete intervals (and thus recalculates $\Rc(x)$) at a relatively low update rate, DTCBFs suit this setting better than their continuous-time counterparts. The barrier function defines safe and unsafe regions of the configuration space:
\begin{equation}\label{eq:barrier}
    h(x_k)=\sigma_{\text{crit}}(\Rc(x_k))-\sigma_{\min},
\end{equation}
where $\sigma_{\min}>0$ is a safety margin and $k$ is the discrete-time index. The safe set is defined as $\mathcal{S} = \{x \mid h(x) \geq 0\}$, and forward invariance of $\mathcal{S}$ is guaranteed by enforcing
\begin{equation}\label{eq:bf_constraint}
    h(x_{k+1}) \geq (1-\alpha)h(x_k),
\end{equation}
where $\alpha\in(0,1)$ sets the allowable decay rate, with larger $\alpha$ permitting more aggressive motion toward the boundary \cite{agrawal2017discrete}. By \cite{agrawal2017discrete}, satisfaction of Eq.~\ref{eq:bf_constraint} at each timestep is sufficient to guarantee forward invariance of the safe set, provided $h(x_0) \geq 0$ at initialization. Since \cite{usevitch2025triangledecomposable} establishes $\sigma_{\text{crit}}>0$ throughout the rigid workspace, we select $\sigma_{\min}$ empirically and verify $h(x_0)\geq0$. Using the discrete-time kinematics,
\begin{equation}
x_{k+1}=x_k+\Delta t\,\dot{x}_k,
\end{equation}
the barrier constraint becomes a function of $\dot{x}_k$ and is incorporated directly into Eq.~\ref{eq:qp_new} as an inequality constraint, ensuring rigidity is preserved at each control step.

This formulation offers three benefits: 1) mathematical guarantees of rigidity preservation, 2) near-maximal workspace utilization by allowing operation close to the boundary, and 3) smooth corrective behavior near singular configurations. These guarantees are contingent on the kinematic model's accuracy in computing $x_{k+1}$, which we assume in this work to be sufficiently high for the guarantees to hold in practice. As discussed in Section~\ref{sec:discussion}, unmodeled effects such as tube compliance can degrade this guarantee; characterizing and compensating for these effects is left to future work.

\subsection{Unified Augmented Optimization Problem}
The final augmented optimization problem solved at each timestep $k$ is formulated as:
\begin{equation}\label{eq:qp_new}
\begin{alignedat}{3}
\min_{\dot{x}_k} \; & \|\Rc\dot{x}_k\|^2 + k_f (\Delta L_k)^\top \Rc \dot{x}_k \\
\text{s.t.} \; 
& A_{\text{move}}\dot{x}_k = b_{\text{move}}, &\quad& \text{motion constraint} \\
& A_{\text{fixed}}\dot{x}_k = 0,              &\quad& \text{fixed nodes} \\
& A_{\text{loop}}\dot{x}_k = 0,               &\quad& \text{loop closure} \\
& A_{\text{broken}}\dot{x}_k = 0,             &\quad& \text{broken elements} \\
& h(x_k + \Delta t \cdot \dot{x}_k) \ge (1-\alpha) h(x_k) &\quad& \text{safety constraint}
\end{alignedat}
\end{equation}
This nonlinear program is solved using Sequential Least Squares Programming (SLSQP) as implemented in SciPy~\cite{scipy,slsqp_kraft}, ensuring real-time performance while satisfying the nonlinear safety constraints. 
Unlike classical Jacobian pseudoinverse or null-space projection methods, this formulation enables the direct incorporation of inequality constraints, such as the model-based rigidity-preserving constraint in Eq.~\ref{eq:bf_constraint}.

\subsection{Closed-Loop Control with Forward Kinematics Estimator} \label{sec:fk}

Given the robot's current configuration, the augmented optimization problem solves for the optimal vertex velocities that produce the desired target vertex motion. We build on this open-loop control by incorporating closed-loop feedback from onboard motor encoders, solving the optimizer at each timestep using the latest state estimate.

\begin{figure*}
\centering
\includegraphics[width=.94\linewidth]{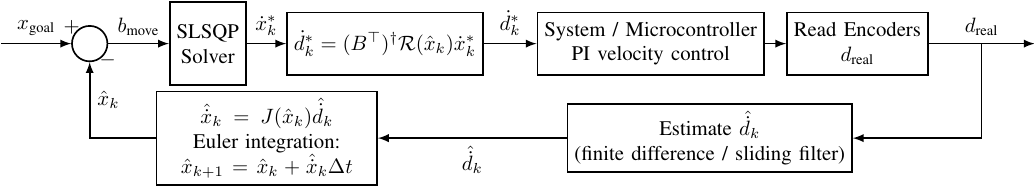}
\caption{Augmented closed-loop kinematic control diagram. The SLSQP solver computes optimal vertex velocities $\dot{x}_k^*$, converted into roller commands for microcontroller PI velocity control. Encoder readings $d_{\text{real}}$ are used to estimate $\hat{\dot{d}}_k$, which is mapped through the kinematic Jacobian to obtain $\hat{\dot{x}}_k$, then propagated via Euler integration to produce feedback $\hat{x}_k$ for the solver.\vspace{-3ex}}
\label{fig:control_diagram}
\end{figure*}

Fig.~\ref{fig:control_diagram} shows the isoperimetric robot's closed-loop controller. The term $b_{\text{move}} \in \mathbb{R}^{D}$ encodes the positional error, calculated by subtracting the target vertex's current position from the goal position. A central computer runs SLSQP to solve Eq.~\ref{eq:qp_new} for a $\dot{x}_k^*$ that moves the target vertex in the direction of $b_{\text{move}}$. This $\dot{x}_k^*$ is converted into roller velocities through Eq.~\ref{eq:IK} and transmitted to onboard microcontrollers via radio modules. Prior work controlled the isoperimetric truss in an open-loop fashion, estimating the updated robot configuration directly from $\dot{x}_k^*$ without encoder feedback \cite{14_usevitch_untethered}.

This paper extends the control process by reading and transmitting motor encoder values from each roller module to the central computer. These values are used to estimate the previous timestep's average roller velocities $\hat{\dot{d}}_k$ via finite differencing (or a sliding low-pass filter under significant encoder noise), which are then propagated through the forward kinematics to update the robot's state estimate. Combining Eqs. \ref{eq:rigidity} and \ref{eq:Ldot_rollers}, roller and vertex velocities are coupled through the relation in Eq.~\ref{eq:jacobian}. As shown in \cite{14_usevitch_untethered}, appending $A_\text{fixed}$ to $\Rc(x)$ removes the rigid-body modes, yielding a full-rank square matrix that can be inverted for the forward kinematics Jacobian
\begin{equation}\label{eq:jacobian}
 \dot{x}= \begin{bmatrix} 
 \Rc(x) \\
 A_\text{fixed}   
 \end{bmatrix}^{-1} 
 \begin{bmatrix} 
 B^\top  \\ 
 0
 \end{bmatrix}
 \dot{d}= J(x)\dot{d}.
\end{equation}

Vertex positions are updated via forward Euler integration with fixed substepping. Since $J(x)$ depends on the configuration through $\Rc(x)$, each substep advances the roller positions, recomputes $J(x)$, and updates the vertex positions. This iterative refinement improves accuracy when the Jacobian varies significantly over a timestep, as expected given the robot's low control frequency. Although higher-order methods such as RK4 could further improve accuracy, the small substep size sufficiently limits integration error.

\subsection{Experimental Platform and Validation Setup} \label{sec:2D_testbed}
To validate our results, we constructed a two-dimensional isoperimetric robot composed of three triangles (Fig.~\ref{fig:2D_robot_diagram}). The testbed rests horizontally on a flat surface with the truss plane parallel to the ground. To remove the planar rigid-body modes without over-constraining the structure, we fixed one corner (V1, Fig.~\ref{fig:2D_robot_diagram}b) in both planar directions and another (V3) in the Y-direction only. Ground-truth motion of the target vertex is recorded with an HTC Vive tracker (SteamVR lighthouse tracking, 4 base stations) and logged at approximately 1 kHz. Motion capture data is used solely for offline evaluation and is not available to the controller. The controller operates at 2\,Hz, with the substepping scheme of Section~\ref{sec:fk} compensating for the resulting inter-sample configuration changes. In addition to this 2D hardware testbed (Fig.~\ref{fig:2D_robot_diagram}), we validate our workspace and manipulability results using 2D and 3D Python simulations. 

\section{Results}

\subsection{Modeling Degraded Kinematics Due to Roller Failure}

\begin{figure}
    \centering
    \includegraphics[width=\linewidth]{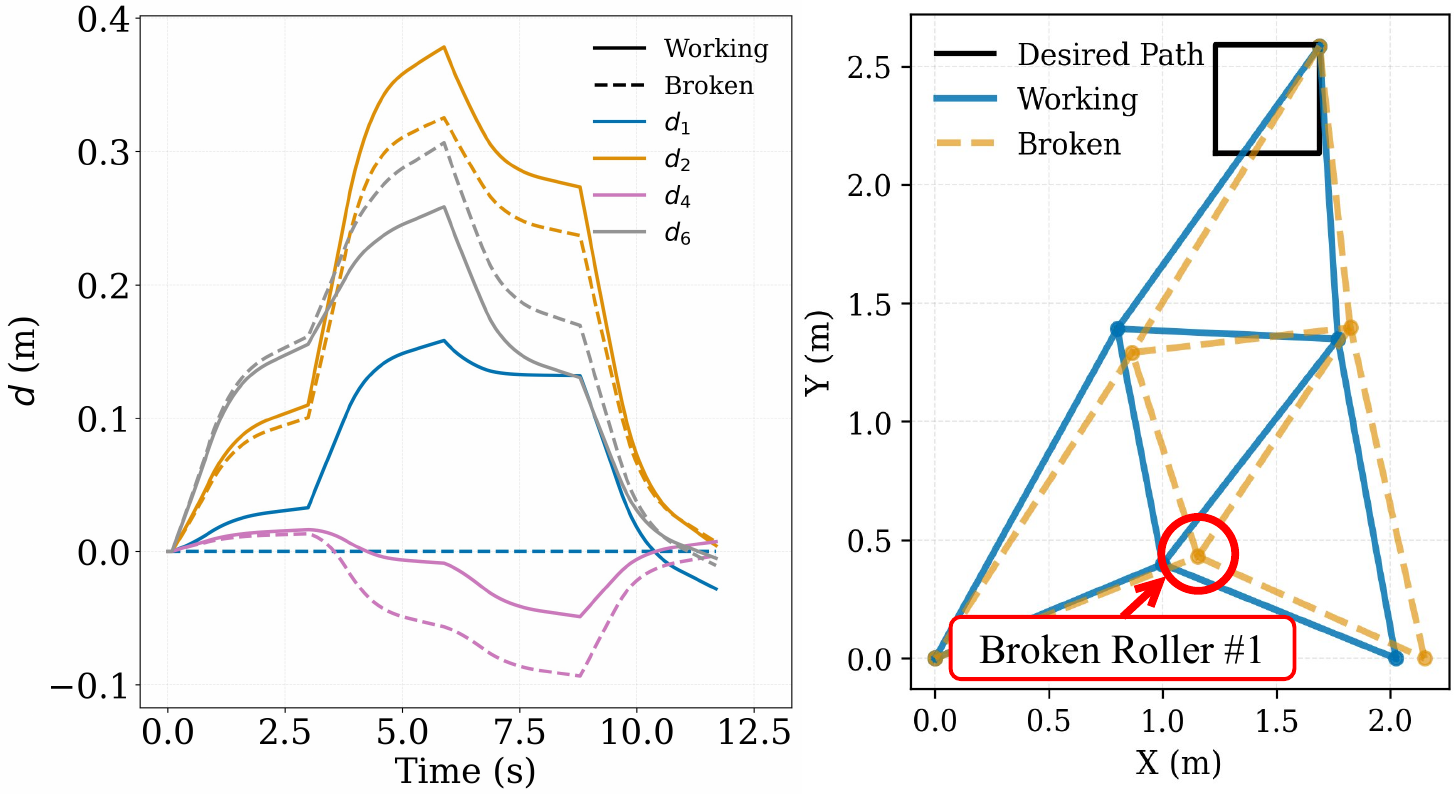}
    \caption{Effect of constraining Roller 1 to zero displacement via Eq.~\ref{eq:broken}. (Left) Roller positions over time for the nominal all-rollers-on [solid] and Roller \#1 broken [dashed] cases. In the broken case, Roller 1 [blue] is fixed at zero while other rollers compensate. (Right) Overlay of nominal and broken configurations at the same target vertex position, showing identical square trajectories despite differing internal geometry.\vspace{-2ex}}
    \label{fig:broken_roller_comparison}
\end{figure}

To simulate roller failures, we apply the constraint in Eq.~\ref{eq:broken}, which fixes a roller's position along its tube. Fig.~\ref{fig:broken_roller_comparison} (left) compares the nominal case, with all rollers functional, against a degraded scenario with Roller 1 constrained to zero motion, both tracing a square path. Fig.~\ref{fig:broken_roller_comparison} (right) shows the effect of this optimization on the robot's configuration while still tracing the same path. This framework extends to any number of rollers, enabling simulation of arbitrary failure combinations.

\begin{figure*}[tbp]
    \centering
    \includegraphics[width=.94\linewidth]{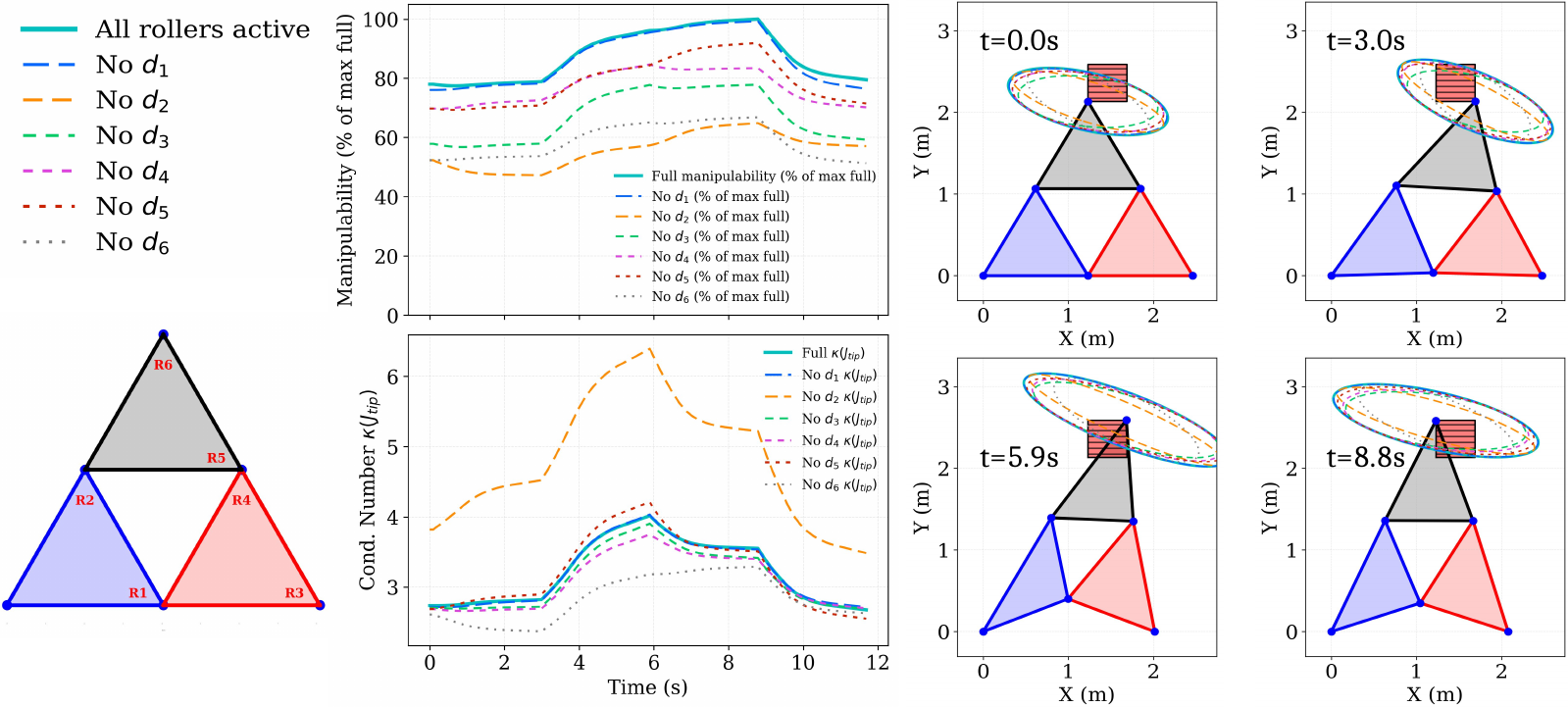}
    \caption{Manipulability analysis of the target vertex as the 2D robot traces a square path. The bottom-left diagram shows the location of each roller module within the 2D truss. The center plot shows the manipulability retention percentage when each individual roller fails, normalized by the maximum manipulability achieved during this trajectory. The right snapshots visualize the corresponding manipulability ellipses at four time points, with the cyan ellipse representing full manipulability and colored ellipses showing degraded manipulability for each failed roller.\vspace{-2ex}}
    \label{fig:manipulability_big_figure}
\end{figure*}

We quantify the impact of roller failures through manipulability analysis at the target vertex. From the full Jacobian $J(x)$ in Eq.~\ref{eq:jacobian}, we extract the target vertex rows to obtain the task-space Jacobian $J_t(x)$ and compute
\begin{equation} \label{eq:manipulability}
M = \sqrt{\det(J_tJ_t^\top)}
\end{equation}
\cite{yoshikawa1985manipulability}. In 2D, the manipulability metric is proportional to the area of the ellipse shown in Fig.~\ref{fig:manipulability_big_figure}. Its anisotropy at the target vertex is quantified by the condition number of $J_t$, defined as the ratio of its largest to smallest singular values \cite{salisbury1982articulated}.

To analyze failure scenarios, we exploit the Jacobian's structure (Eq.~\ref{eq:jacobian}): each column represents an individual roller's contribution to task-space motion. Zeroing the $i$th column yields a degraded Jacobian modeling the instantaneous effect of the $i$th roller failing at that configuration. Computed for each roller at every time step, this degraded metric reveals each roller's importance to the robot's motion capability throughout the trajectory. Fig.~\ref{fig:manipulability_big_figure} presents this analysis: the time evolution of manipulability under individual roller failures (normalized by the trajectory's maximum value), along with manipulability ellipses at key time points, offering an intuitive geometric view of how each failure degrades directional motion capability. The manipulability index grows as the robot extends outward, reflecting the kinematic structure's combined revolute- and prismatic-type motion. However, a large index alone can be misleading, so we also report the Jacobian's condition number to characterize the ellipse's anisotropy, which worsens as the robot extends further from its neutral configuration.

\subsection{Degradation of Workspace due to Roller Failure}
\begin{figure*}
    \centering
    \includegraphics[width=0.95\textwidth]{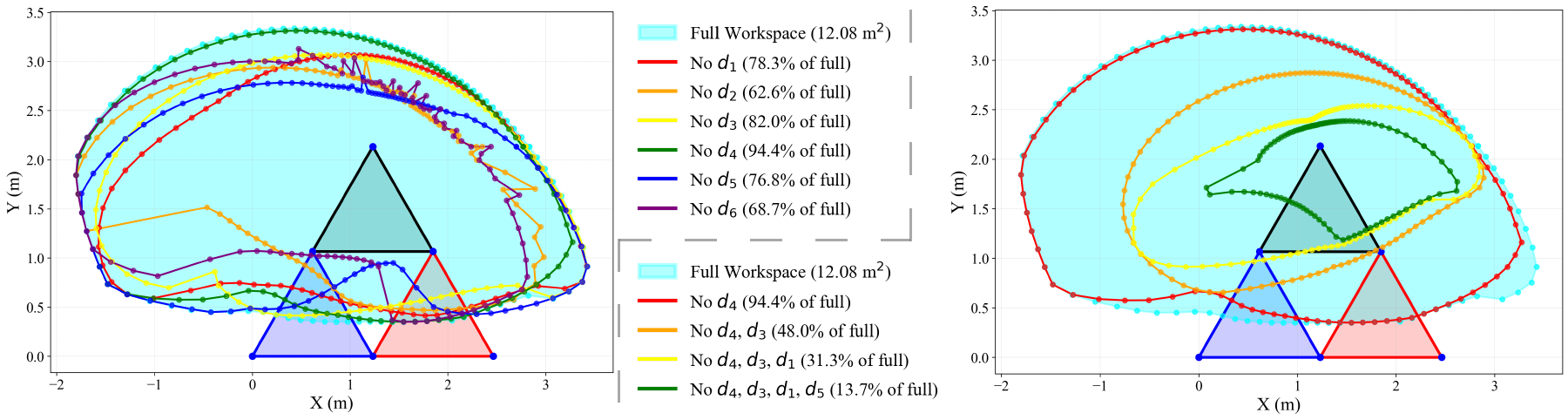}
    \caption{Workspace analysis of the truss robot, showing the effective workspace of the robot with individual roller units or a combination of roller units deactivated. (Left) The effect of individual roller unit failure on the robot's workspace. Ranking rollers by individual failure impact, from least to most severe, gives the order 4, 3, 1, 5, 6, 2. (Right) The effect of combined roller unit failure on the robot's workspace, with rollers deactivated cumulatively in the order established in Fig.~\ref{fig:workspace_cbf} (left).\vspace{-2ex}}
    \label{fig:workspace_cbf}
\end{figure*}

Fig.~\ref{fig:workspace_cbf} shows the effect of roller failures on the robot's workspace, computed by commanding the target vertex to extend radially in uniformly distributed directions until the optimizer can no longer find a feasible solution \cite{usevitch2025triangledecomposable}. Individual roller failures produce asymmetric workspace reductions, with some rollers having substantially greater impact than others; however, single roller failures preserve greater than 62\% of the nominal 2D workspace area, demonstrating the robot's inherent robustness to individual actuator loss. Following the impact-ranked order (4, 3, 1, 5, 6, 2), Fig.~\ref{fig:workspace_cbf} (right) shows the reduction of feasible workspace under combined failures.

\subsection{Rigidity Maintenance with DTCBF Constraints}
\begin{figure*}[p]
    \centering
    \includegraphics[width=\linewidth]{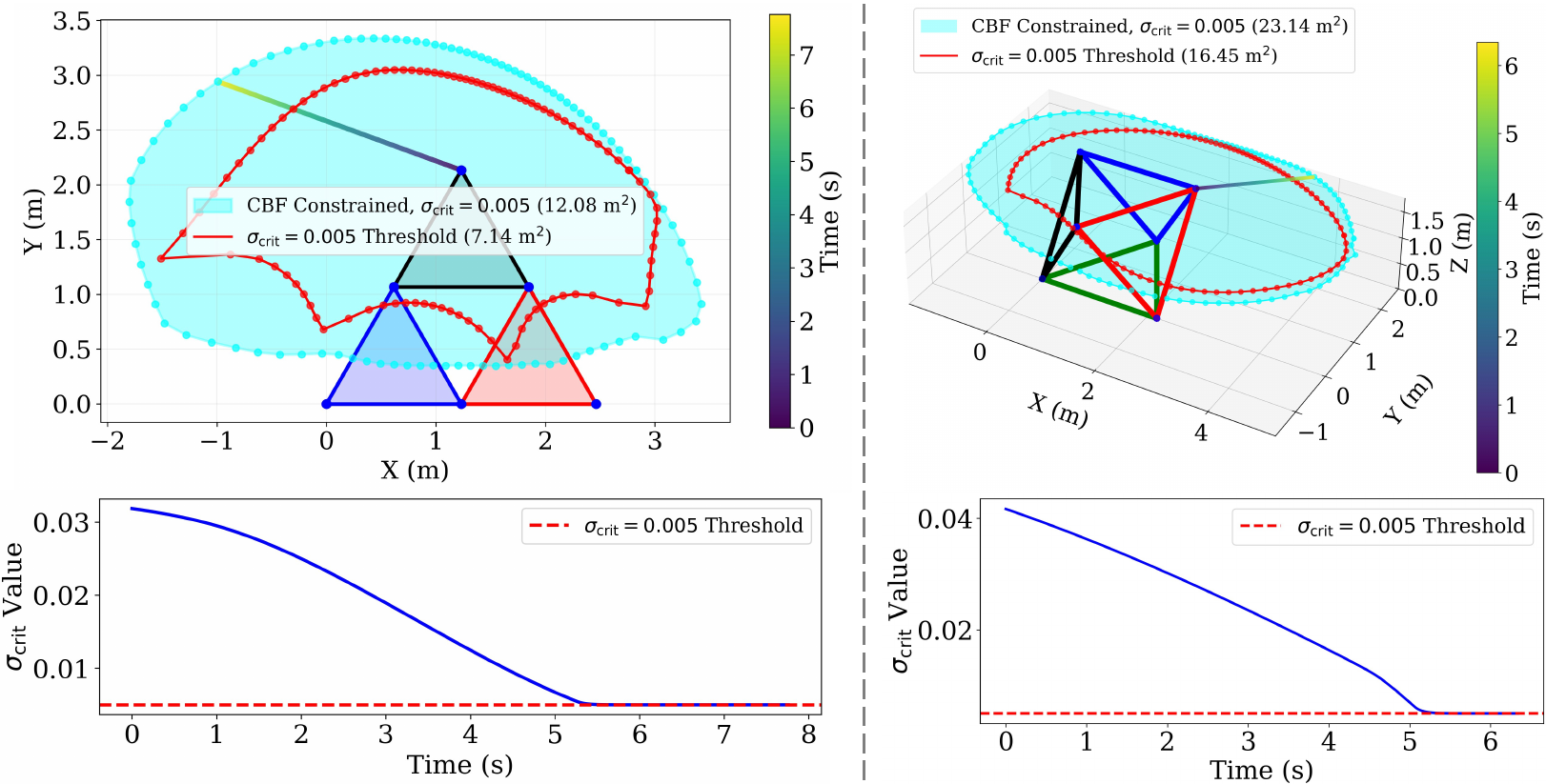}
    \caption{Reachable workspace under the DTCBF constraint for the 2D (left) and 3D (right) truss configurations. (Top left) 2D workspace comparison: DTCBF-constrained reachable area (cyan, 12.08~m$^2$) versus the fixed $\sigma_\text{min} = 0.005$ threshold boundary (red, 7.14~m$^2$), a 69\% increase in reachable area. (Top right) Horizontal-plane workspace of the target vertex for a 3D octahedral module: DTCBF-constrained reachable area (cyan, 23.14~m$^2$) versus the fixed-threshold boundary (red, 16.45~m$^2$), a 41\% increase. In both cases, the DTCBF expands the workspace by readjusting the robot configuration to satisfy the barrier constraint rather than enforcing a hard $\sigma_\text{crit}$ cutoff, demonstrating that the approach also applies to the 3D formulation. Each example trajectory is color-coded by elapsed time. (Bottom) Evolution of $\sigma_\text{crit}$ along the corresponding trajectory for the 2D (left) and 3D (right) cases, each converging to the threshold near $t = 5$~s.}

    \label{fig:barrier_workspace}
\end{figure*}

Fig.~\ref{fig:barrier_workspace} compares the robot's workspace with and without the DTCBF, for both the 2D testbed configuration and a 3D octahedral module. Without the DTCBF, the robot halts once $\sigma_\text{crit}$ reaches the fixed threshold $\sigma_\text{min}$. The DTCBF increases the reachable workspace by 69\% in the 2D case and by 41\% in the 3D case, demonstrating that it can exploit null-space motion rather than enforcing a hard rigidity cutoff, and that this benefit is not specific to the 2D formulation. Fig.~\ref{fig:barrier_workspace} also shows that $\sigma_\text{crit}$ remains within the safe region defined by $h(x)$ throughout each example trajectory, color-coded by time, for both configurations. Thus, unlike previous methods that only maintain rigidity \cite{usevitch2025triangledecomposable}, the proposed DTCBF provides a model-based guarantee of rigidity while also enlarging the workspace by reconfiguring the robot within the Jacobian null space.

\subsection{Effect of Closed-Loop Control on Positional Accuracy}

\begin{figure*}[p]
    \centering
    \includegraphics[width=0.95\linewidth]{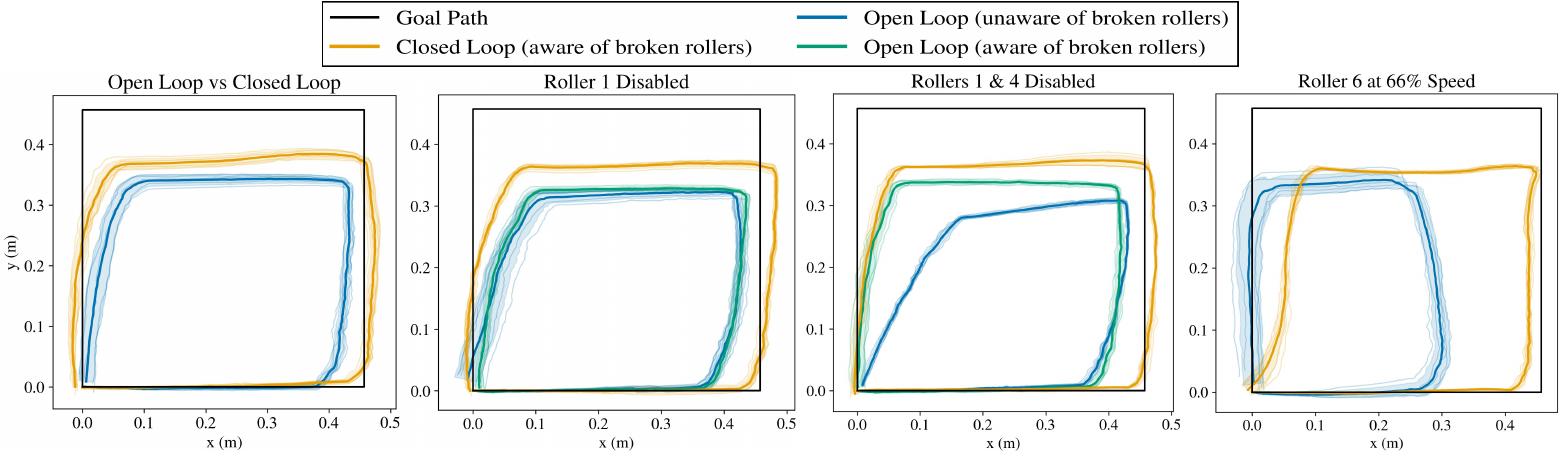}
    \caption{Comparison of open-loop and closed-loop control on the physical robot tracing a 0.45 m $\times$ 0.45 m square across various roller module operational scenarios, each repeated 5 times, with shaded regions showing $\pm 1$ standard deviation from the mean.}
    \label{fig:physical_open_vs_closed}
\end{figure*}

Fig.~\ref{fig:physical_open_vs_closed} compares open-loop (OL) and closed-loop (CL) performance across operational scenarios, with the target vertex tracing a square path by visiting each corner sequentially. For complete roller failures, we distinguish failure-unaware (OL-U) from failure-aware open-loop control (OL-A). Table~\ref{tab:rmse} reports each trajectory's root-mean-square error (RMSE) relative to the desired path, along with the performance improvements over open-loop control. The optimization solves in $27$\,ms (median; mean $33$\,ms; 99th percentile $120$\,ms), typically well within the $500$\,ms control period. The solver's tunable 100-iteration cap was reached in rare cases, producing worst-case solve times of ${\sim}750$\,ms. When this occurred, the controller intentionally halted the robot until the next motion command.

The improvements in Table~\ref{tab:rmse} stem from two sources: encoder-based positional feedback and the optimizer's explicit failure awareness. For scenarios in which the actuator condition is fully known, CL consistently outperforms OL, improving RMSE by $>22\%$ relative to the fully functional OL baseline, showing that closed-loop control on a partially functioning robot can outperform open-loop control with all rollers operational. Additionally, under each actuator-failure condition, CL reduces RMSE by $>34\%$ relative to OL-U. OL-A also outperforms OL-U, and the OL-A RMSE values for the broken scenarios remain close to the fully functional OL baseline, indicating that the constraints in Eq.~\ref{eq:qp_new} redistribute actuation to maintain tracking despite the reduced actuator set. In contrast, the final scenario considers unmodeled partial roller degradation, with Roller~6 operating at $66\%$ of its commanded speed. Such degradation cannot be encoded in $\mathcal{F}$, which admits only complete failures, so the controller is necessarily unaware of it. Despite this mismatch, CL control improves RMSE by $39.2\%$ over OL-U and achieves $7.1\%$ lower RMSE than the fully functional OL baseline, demonstrating robustness to actuator-model mismatch through feedback. Endpoint repeatability is generally better under CL, though with only five repetitions per scenario we do not draw conclusions from these differences. RMSE standard deviation shows no consistent CL advantage, suggesting that trajectory-shape repeatability is dominated by unmodeled effects like friction, structural compliance, and velocity-tracking errors.

\begin{table*}[p]
\centering
\caption{Average trajectory tracking performance, each scenario repeated 5 times. RMSE is the root-mean-square error of the traced path relative to the goal path, and endpoint standard deviation is the standard deviation of the robot's final position across repetitions. Global improvement is measured relative to the fully functional OL baseline; local improvement is measured relative to OL Unaware (OL-U) within each failure condition, quantifying the benefit of failure-aware control.}
\label{tab:rmse}
\resizebox{2\columnwidth}{!}{
\begin{tabular}{lcccccccccc}
\toprule
 & \multicolumn{2}{c}{\textbf{All Rollers On}} & \multicolumn{3}{c}{\textbf{Broken 1}} & \multicolumn{3}{c}{\textbf{Broken 1 \& 4}} & \multicolumn{2}{c}{\textbf{Roller 6 at 66\% Speed}} \\
\cmidrule(lr){2-3} \cmidrule(lr){4-6} \cmidrule(lr){7-9} \cmidrule(lr){10-11}
 & OL & CL & OL-U (Unaware) & OL-A (Aware) & CL & OL-U (Unaware) & OL-A (Aware) & CL & OL-U (Unaware) & CL \\
\midrule
RMSE Mean (mm) & 56.7 & 39.3 & 67.6 & 63.4 & 44.1 & 80.4 & 59.7 & 43.1 & 86.7 & 52.7 \\
RMSE Std Dev (mm) & 2.8 & 2.9 & 2.9 & 1.2 & 1.5 & 1.1 & 0.9 & 1.8 & 0.5 & 1.2 \\
Endpoint Std Dev (mm) & 11.6 & 4.0 & 21.2 & 6.7 & 2.3 & 3.2 & 9.7 & 5.4 & 16.0 & 11.7 \\
Improvement Global (\%) & -- & 30.7 & -19.2 & -11.8 & 22.3 & -41.7 & -5.3 & 24.0 & -52.9 & 7.1 \\
Improvement Local (\%) & -- & 30.7 & -- & 6.3 & 34.9 & -- & 25.7 & 46.4 & -- & 39.2 \\
\bottomrule
\end{tabular}
}
\end{table*}

\section{Discussion}\label{sec:discussion}
These analyses collectively inform users of the robot's capabilities and operational considerations. Triangular modules allow assembly in many configurations for the same task; however, the risk of roller failure should guide users toward placing more reliable hardware at critical locations, since reliability can diverge over time through normal operation. Forecasting which motor failure would have the greatest effect on task performance is therefore essential for assembly and deployment. Since the manipulability index, ellipses, condition number, and workspace analysis are general metrics, the analyses in Figs.~\ref{fig:manipulability_big_figure} and~\ref{fig:workspace_cbf} apply to arbitrary robot configurations. Workspace and manipulability are also sensitive to \textit{where} a failure occurs in the trajectory, and the same tools can analyze degraded configurations at any operational point. These tools can also identify active rollers that contribute minimally to a task and could be replaced with passive rollers without sacrificing performance, informing cost-effective robot design.

The trajectories in Fig.~\ref{fig:physical_open_vs_closed} show imperfect motion and increased error in the Y-direction across all scenarios. Although closed-loop control extends the trajectory toward the desired path in the X-direction, its Y-direction extension remains largely unchanged from open-loop control. We attribute this to unmodeled bowing in the inflated tubes, visible in the upper triangle of Fig.~\ref{fig:system_diagram}. This bowing is a result of external loading and of plastic deformation in the tube material accumulated from repeated use and prolonged loading of the same tubing during earlier testing \cite{stanciu2026modular}. Because the closed-loop controller estimates the robot's configuration via forward kinematics under an idealized rigid model, it cannot account for tube compliance; encoder feedback only measures roller displacement, not deviation from the rigid-body assumption. Future work will characterize this deformation and incorporate it into the closed-loop controller, enabling meaningful hardware validation of the DTCBF's workspace-expansion boundary. Trajectory tracking could also be improved by penalizing perpendicular distance from the desired path in addition to distance to the goal, particularly under imperfect velocity control. These and other trajectory-planning improvements are left for future work.

\section{Conclusion}
Continued advancements in truss robot design could pave the way for soft, inflatable alternatives to rigid, heavy, metallic structures. This work enhanced the robustness of isoperimetric robots in three ways. First, we introduced a controller that allows the robot to operate without a subset of its active roller modules, quantifying the effect of roller failures on manipulability and reachable workspace. Second, we implemented a DTCBF to help the robot avoid singular configurations. Third, we implemented closed-loop control, improving robustness to rollers that fail to achieve their commanded speeds. These improvements contribute to more robust, lightweight, and reconfigurable robotic truss systems for operation in challenging environments.

\bibliographystyle{bibtex/bst/IEEEtran}
\bibliography{bibtex/bib/bibliography}

\end{document}